\documentclass[a4paper,twoside]{article}

\usepackage{epsfig}
\usepackage{subcaption}
\usepackage{calc}
\usepackage{amssymb}
\usepackage{amstext}
\usepackage{amsmath}
\usepackage{amsthm}
\usepackage{multicol}
\usepackage{multirow}
\usepackage{graphicx}
\usepackage[hidelinks]{hyperref}
\usepackage[utf8]{inputenc}
\usepackage{textcomp}
\usepackage{natbib}

\usepackage{pslatex}
\usepackage{apalike}
\usepackage{algorithm2e}
\usepackage[bottom]{footmisc}
\usepackage{xcolor}
\usepackage{booktabs}

\usepackage{SCITEPRESS}     

\begin{document}

\title{Benchmarking and Enhancing PPG-Based Cuffless Blood Pressure Estimation Methods}

\author{\authorname{Neville Mathew\sup{1}\orcidAuthor{0009-0000-3213-364X}, Yidan Shen\sup{2}\orcidAuthor{0009-0005-8412-7854}, Renjie Hu\sup{3}\orcidAuthor{0000-0002-0496-6035}, Maham Rahimi\sup{4}, and George Zouridakis\sup{1,2,3,5}\orcidAuthor{0000-0002-7770-9857}}
\affiliation{\sup{1}Department of Engineering Technology, University of Houston, Texas, USA}
\affiliation{\sup{2}Department of Electrical and Computer Engineering, University of Houston, Texas, USA}
\affiliation{\sup{3}Department of Information Science Technology, University of Houston, Texas, USA}
\affiliation{\sup{4}Department of Cardiovascular Surgery, Houston Methodist Hospital, Texas, USA}
\affiliation{\sup{5}Department of Biomedical Engineering, University of Houston, Texas, USA}
\email{ mrahimi@houstonmethodist.org, \{namathew3, yshen20, rhu7, zouridakis\}@uh.edu}
}


\abstract{
Cuffless blood pressure screening based on easily acquired photoplethysmography (PPG) signals offers a practical pathway toward scalable cardiovascular health assessment. Despite rapid progress, existing PPG-based blood pressure estimation models have not consistently achieved the established clinical numerical limits such as AAMI/ISO 81060-2, and prior evaluations often lack the rigorous experimental controls necessary for valid clinical assessment. Moreover, the publicly available datasets commonly used are heterogeneous and lack physiologically controlled conditions for fair benchmarking. To enable fair benchmarking under physiologically controlled conditions, we created a standardized benchmarking subset NBPDB comprising 101,453 high-quality PPG segments from 1,103 healthy adults, derived from MIMIC-III and VitalDB. Using this dataset, we systematically benchmarked several state-of-the-art PPG-based models. The results showed that none of the evaluated models met the AAMI/ISO 81060-2 accuracy requirements (mean error $<$ 5 mmHg and standard deviation $<$ 8 mmHg). To improve model accuracy, we modified these models and added patient demographic data such as age, sex, and body mass index as additional inputs. Our modifications consistently improved performance across all models. In particular, the MInception model reduced error by 23\% after adding the demographic data and yielded mean absolute errors of 4.75 mmHg (SBP) and 2.90 mmHg (DBP), achieves accuracy comparable to the numerical limits defined by AAMI/ISO accuracy standards. Our results show that existing PPG-based BP estimation models lack clinical practicality under standardized conditions, while incorporating demographic information markedly improves their accuracy and physiological validity.
}

\onecolumn \maketitle \normalsize \setcounter{footnote}{0} \vfill

\section{\uppercase{Introduction}}
\label{sec:introduction}

Blood pressure (BP) is a critical physiological indicator of cardiovascular health. Accurate assessment of systolic and diastolic blood pressure (SBP and DBP) is essential for diagnosing and managing hypertension, a leading cause of global morbidity and mortality \citep{flint2019effect, magder2018meaning}. Cuffless BP screening could revolutionize preventive care and enable large-scale population health assessment. However, conventional cuff-based devices—such as mercury and oscillometric sphygmomanometers—are cumbersome for frequent or daily use, as they require manual operation and disrupt normal activities \citep{tale2021sphygmomanometers}.

Recent progress in wearable sensing technologies has revitalized research on non-invasive and cuffless blood pressure (BP) estimation. In particular, optical sensing approaches, such as those used in pulse oximetry \citep{elgendi2019photoplethysmography}, have been extensively investigated for their potential in cuffless BP screening. Among available physiological signals used for cuffless BP estimation, photoplethysmography (PPG) has emerged as an attractive marker due to its simplicity and compatibility with wrist or fingertip-based devices as well as cost-effective equipment requirements. Based on the Beer–Lambert law, PPG captures pulsatile changes in micro-vascular blood volume, indirectly reflecting hemodynamic changes \citep{elgendi2019photoplethysmography, chen2024ppg}. However, PPG measurements are susceptible to motion artifacts \citep{yan2005reduction} and variations in skin pigmentation \citep{cabanas2022skin}, thereby demanding robust algorithms and careful data quality control.

Machine learning (ML) and deep learning (DL) approaches have shown significant promise for estimating BP from PPG waveforms. Early studies extracted hand-crafted morphological and temporal features \citep{chu2023noninvasive}, while more recent work has shifted toward end-to-end deep models such as one-dimensional convolutional networks (1D-CNNs) and sequence models like S4 \citep{ELHAJJ2021102301, gu2022efficientlymodelinglongsequences, moulaeifard2025generalizabledeeplearningphotoplethysmographybased}. 
Despite these advances, no systematic benchmarking has been conducted to assess whether existing models meet the numerical accuracy requirements of the AAMI/ISO 81060-2 standard (mean error $<5$ mmHg and standard deviation $<8$ mmHg). Most studies report statistical metrics like MAE or correlation on heterogeneous datasets, without evaluating clinical compliance or reproducibility across conditions. As a result, it remains unclear whether these data-driven models achieve clinically acceptable accuracy. Establishing a benchmarking framework that aligns model evaluation with medical device standards is therefore essential to bridge the gap between algorithmic progress and clinical applicability.

Several studies have explored multimodal strategies that combine PPG with complementary physiological signals such as electrocardiogram (ECG) or pulse transit time (PTT) \citep{escobarpaper, nie2024reviewdeeplearningmethods,raza2025neuromoetransformerbasedmixtureofexpertsframework}. These approaches can improve estimation accuracy by leveraging additional temporal or hemodynamic information. However, collecting ECG or PTT data typically requires specialized sensors and controlled experimental conditions, which limits their practicality for continuous or large-scale screening. Given that our primary objective is to assess model performance and physiological relevance using widely available, low-cost PPG data, frameworks incorporating multi-physiological signal inputs are beyond the scope of the present study.

Another challenge for the PPG based BP estimation models lies in the nature of existing publicly available datasets, which often comprise heterogeneous clinical cohorts including patients with comorbidities, unstable hemodynamics, or hypertension. Such uncontrolled heterogeneity obscures the physiological mechanisms that govern the relationship between PPG and BP, making it difficult to disentangle algorithmic performance from population effects. Consequently, models trained on mixed populations may inadvertently exploit pathological or disease-specific patterns correlated with BP, rather than learning the underlying physiological principles of cardiovascular regulation \citep{mehta2024examining, yusheng}. As a result, it remains unclear how the performance of existing PPG-based models would compare under standardized and physiologically controlled conditions.

Another critical yet often overlooked factor in existing cuffless BP studies is the role of demographic variables. Physiological determinants such as age, sex, and body mass index (BMI) strongly influence vascular compliance and peripheral resistance \citep{Kohn2015, evans_bmi}, thereby modulating the PPG–BP relationship. Ignoring these variables not only limits model interpretability, but also biases evaluation by conflating algorithmic effects with demographic composition.

To address these limitations, we first establish a systematic benchmarking framework to evaluate state-of-the-art PPG-based BP estimation models under well-controlled, physiologically standardized conditions and determine whether they achieve clinically acceptable accuracy. We then extend these models by explicitly incorporating demographic information—age, sex, and BMI—into their design, and re-evaluate their performance to quantify the contribution of demographic factors to model accuracy and generalization.

To support this benchmarking effort, we introduce NBPDB (Normal Blood Pressure Database), a standardized benchmarking subset systematically filtered from publicly available VitalDB and MIMIC-III databases for cuff-less, PPG-based BP estimation in healthy adults \citep{HU20239}. NBPDB represents the controlled scenario defined in our framework, focusing on subjects with normal systolic and diastolic ranges and stable hemodynamic conditions. Leveraging NBPDB, we systematically evaluate multiple state-of-the-art neural architectures under both calibration-based and calibration-free settings and further extend them into demography-aware variants to quantify the contribution of age, sex, and BMI to model accuracy and generalization.

The remainder of this paper is organized as follows. Section 2 provides an in-depth review of existing literature on cuffless, PPG-based blood pressure estimation and related machine learning approaches. Section 3 details the construction of the NBPDB dataset and describes how existing state-of-the-art models are extended into a demography-aware framework. Section 4 presents the benchmarking results and compares model performance under both calibration-based and calibration-free settings. Section 5 discusses the limitations and potential directions for future improvement, and section 6 concludes the paper.

\section{\uppercase{Related works}}

\subsection{BP Prediction}

Cuffless blood pressure estimation is dominated by two methodological categories: (i) PPG augmented with additional physiological signals, which enhances stability but requires multiple sensors, and (ii) PPG-only models, which optimize portability but with reduced prediction accuracy.

\textbf{PPG with additional physiology signals}
Early systems fused PPG with timing cues such as PAT, PTT or time-to-peak (TTP) \citep{pat_paper, pwv_paper}, often with linear or polynomial regression. These pipelines work in controlled settings but depend on reliable R-peak detection and precise sensor application; even a small calibration drift or a missing PPG signal can degrade accuracy and limit real-world prediction estimates.

Deep-BP \citep{yan2019novel} introduced a CNN with hard parameter-sharing multi-task learning to jointly estimate SBP and DBP from synchronized ECG+PPG signals, using mean-filter denoising and classic MTL sharing \citep{caruna1993multitask}. \citep{huang2022mlp} adapted MLP-BP for gMLP/LSTM-Mixer-style token mixing to multi-channel ECG+PPG, aiming to replace hand-crafted features. These approaches reduce manual engineering but still rely on multi-channel signal availability and tightly controlled preprocessing; moreover, several reports emphasize calibration or subject-overlapping splits, which can mask the true generalization on unseen subjects.

\textbf{PPG-only}
\citep{sadrawi} formulate genetic deep convolution autoencoders (LeNet and U-Net-based DCAE ensemble models) to reconstruct ABP from single-channel PPG and derive BP from the recovered waveform. The idea stabilizes learning on short segments, but the study uses a small dataset (18 patients) with manual filtering of atypical waveforms—conditions that under-represent real-world noise and motion. 

{\citep{moulaeifard2025generalizabledeeplearningphotoplethysmographybased} retrofit 2D convolution neural networks for PPG signals and emphasize standardized reporting (AAMI/BHS) with validation on external datasets. This improves comparability and cross-dataset awareness. However, the approach overlooks demographic inputs, which can lead to inaccurate blood pressure predictions across individuals across different conditions.

\subsection{BP datasets}

\noindent\textbf{UQVS} \citep{liu2012university}: a dataset collected from a limited number of surgical cases, designed for screening patients under anesthesia with multiple synchronized waveforms. It offers long, continuous Operating Room recordings with reliable synchronization, though the restricted cohort size and operating-room–focused physiology result in relatively narrow demographic and activity coverage.

\noindent\textbf{MIMIC-II/III waveform ecosystem} \citep{goldberger2000physiobank}: a large-scale ICU repository that has become a cornerstone for blood pressure studies, often through different “UCI/MIMIC” subsets. Its breadth and accessibility, with arterial blood pressure serving as a strong reference; however, it is limited by the small number of ICU patient cases available.

\begin{table}[htbp]
    \centering
    \caption{Key Statistics of the Train, CalBased, and CalFree Datasets within NBPDB. SBP and DBP shown with mean$\pm$SD}
    \begin{tabular}{|l|c|c|c|}
        \hline
        \textbf{Metric} & \textbf{Train} & \textbf{CalBased} & \textbf{CalFree} \\
        \hline
        Total Sub&529 & 512 & 62 \\
        \hline
        Total Seg&81088 & 9090 & 11275 \\
        \hline
        Male&43685 & 4890 & 7504 \\
        \hline
        Female&37403 & 4200 & 3771 \\
        \hline
        Age&51.2$\pm$10.9&51.2$\pm$10.9& 49.6$\pm$11.3 \\
        \hline
        BMI&22.0$\pm$1.8&22.0$\pm$1.8&22.2$\pm$1.9\\
        \hline
        SBP&115.3$\pm$8.9&115.3$\pm$8.9&114.9$\pm$8.8\\
        \hline
        DBP&67.9$\pm$5.5&68.0$\pm$5.6& 68.4$\pm$5.5\\
        \hline
    \end{tabular}
    
    \label{tab:dataset_stats}
\vspace{0.1cm}
\end{table}

\noindent\textbf{PulseDB} \citep{pulsedb_paper}: a curated large-scale collection of synchronized PPG/ECG/ABP segments, released with predefined subject-balanced splits supporting both calibration-free and standardized (AAMI/BHS) evaluations. The consistent preprocessing pipeline and well-defined protocols facilitate fair comparisons and external validation, while the highly dynamic hemodynamics typical of ICU cohorts make it challenging to meet clinical accuracy expectations \citep{huang2024validation}.

\noindent\textbf{Small proprietary cohorts} (various, e.g., \citep{sadrawi}): narrowly scoped studies involving tens of patients and substantial manual filtering. Such datasets are valuable for controlled, hypothesis-driven experiments, yet their limited size and diversity constrain the strength of external validation and testing.

\section{\uppercase{Methodology}}

\subsection{Construction of the NBPDB}

To address the noise and heterogeneity inherent in PPG signals and the limitations of existing datasets, this study introduces NBPDB (Normal Blood Pressure Database), a standardized benchmarking subset systematically filtered from two widely used public databases: VitalDB \citep{lee2022vitaldb} and MIMIC-III \citep{johnson2016mimic, physionet}. Both databases contain extensive collections of physiological waveforms—such as ECG, PPG, and arterial blood pressure (ABP) signals—across diverse patient populations and clinical conditions.

However, direct use of these heterogeneous datasets poses two major challenges. First, variability in recording quality and signal characteristics often introduces noise and artifacts, compromising the reliability of downstream analyses. To address this, we constructed NBPDB by applying the standardized data processing pipeline established by PulseDB \citep{pulsedb_paper} to subjects from VitalDB and MIMIC-III. While inheriting PulseDB's rigorous curation and signal-processing framework, NBPDB further refines and filters the original data to construct a clean, standardized benchmarking subset specifically tailored for evaluating blood pressure (BP) prediction models under normal physiological conditions.

Second, while VitalDB and MIMIC-III encompass diverse patient populations, their heterogeneity in age, comorbidities, and hemodynamic states makes it difficult to interpret model performance.
To address this, we further refined NBPDB by selecting samples within clinically defined normal BP ranges: systolic BP (90–120 mmHg), diastolic BP (60–80 mmHg), age (18–65 years) \citep{hardin, ghimire2023geriatric, nih2025}, and BMI (18.5–25 kg/m²) \citep{weir2023bmi, cdc2024bmi}. To accommodate inter-individual variability, we extended acceptable limits to SBP 90–130 mmHg and DBP 60–85 mmHg \citep{whelton}. Subjects lacking BMI data were excluded.

This refined subset establishes a controlled, clinically relevant benchmark that (1) minimizes confounding effects from disease or medication, (2) represents the primary target population for healthy adults, and (3) supports reproducible cross-study comparisons. By ensuring robust signal quality, controlled population selection, and standardized feature extraction, NBPDB provides a reproducible benchmarking framework for fair and interpretable evaluation of BP estimation models under normal physiological conditions.

\begin{figure}[h]
    \centering
    \includegraphics[width=0.499\textwidth]{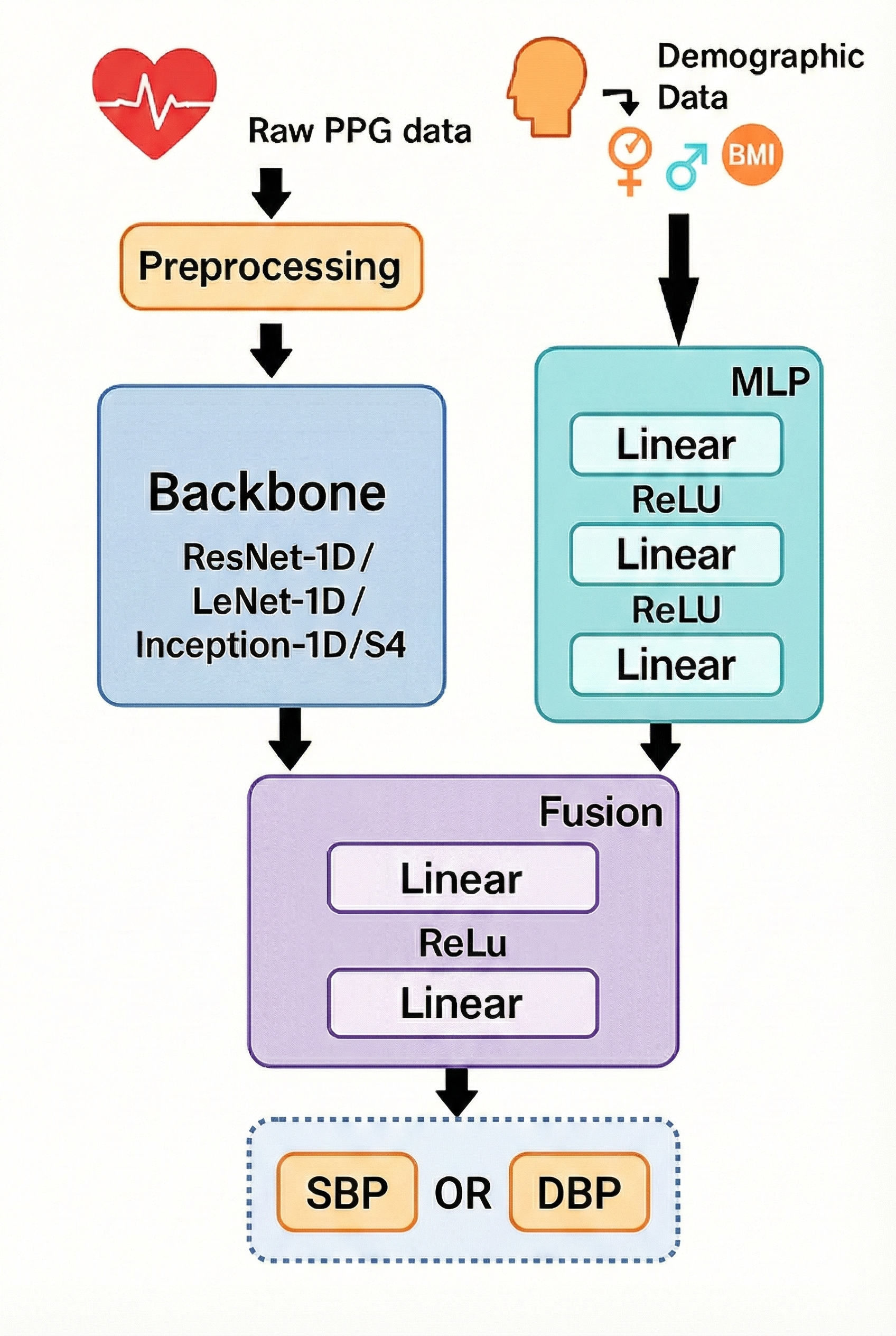}
    \caption{Network architectures. We build five multimodal networks using the same architecture design.}
    \label{fig:network_arch}

\vspace{0.1cm} 
\end{figure}

\subsection{Model Architecture}

In this work, we systematically explore a range of deep learning architectures for blood pressure prediction, leveraging multimodal inputs that include PPG signals and demographic features such as age, sex, and Body Mass Index (BMI). The backbone models evaluated comprised ResNet-1D variants (XResNet-18 and XResNet-50), Inception-1D, LeNet-1D, and Structured State Space Sequence (S4) models—each selected for their unique capabilities in capturing the complex patterns within physiological data. Across all models, a unified design principle is adopted: PPG signals in combination with demographic features are first encoded by dedicated neural networks and combined through a late-fusion module to generate SBP and DBP predictions.

\textbf{Simple Feed-forward CNN Architectures}
Feed-forward CNNs through LeNet-1D \citep{lenet_paper} represents the basic form of convolutional modeling for physiological signals. Unlike ResNet networks, these models have sequential, unidirectional data flow, and the absence of skip connections or residual pathways. In our MLeNet model, the classical LeNet is adapted into a one-dimensional variant (LeNet-1D), which processes raw PPG input through a direct stack of convolution and pooling layers using default hyperparameters. This architecture captures fundamental and localized temporal patterns but does not employ any specific strategies to prevent vanishing gradients or enable information reuse, as seen in more advanced models. In parallel, demographic information is independently encoded using an MLP composed of two layers with 16 hidden units. The features from both modalities are concatenated and processed by a fusion layer with 32 hidden neurons to produce SBP and DBP predictions.

\textbf{ResNet-based Architectures}

Residual Networks (ResNets) \citep{he2015deepresiduallearningimage} have substantially advanced time-series deep learning by introducing skip connections, which promote stable and efficient gradient propagation and enable the construction of deeper, more expressive models. The key principle of ResNet is its solution towards vanishing-gradient issues and allowing to learn temporal representations critical for understanding physiological signals. In our MResNet models, we extend the standard ResNet framework to a multimodal one-dimensional setting for blood pressure prediction. The PPG signals are processed through a one-dimensional XResNet backbone adapted from ResNet-18 and ResNet-50, where we follow the original 1D XResNet configuration but expand the initial convolution layer to 64 channels for enhancing feature capacity of high resolution temporal inputs. Demographic information is also independently encoded using a two-layer MLP, then fused with 1D XResNet backbone by a fusion layer to output SBP and DBP predictions. 

\textbf{Inception-based Architectures}

Inception architectures \citep{Szegedy2016}, originally introduced for computer vision, employ parallel convolution filters of varying kernel sizes within each block, enabling simultaneous extraction of features at multiple temporal scales. This multi-branch mechanism is advantageous for physiological signal analysis, where both localized waveform characteristics and broader temporal dependencies contribute to accurate BP estimation. 

In the MInception model, we employ a one-dimensional Inception backbone specifically adapted for BP prediction from PPG signals. The network comprises of six sequential Inception blocks, each containing parallel convolution branches with kernel sizes of 39, 19 and 9.
Residual connections are inserted every three layers to stabilize the gradients and enhance feature usability. Following the Inception backbone, the global average pooling layer aggregates temporal features into a compact representation, which is fused with demographic features via the same way MLenet or MResNet18/50 performs late-fusion. This structure enables the model to learn both PPG-derived temporal features and static demographic attributes, ensuring comprehensive modeling of individual cardiovascular dynamics.

\textbf{S4-based Models}
Structured State Space Sequence (S4) \citep{Gu2022} models offer a powerful framework for time-series modeling, particularly effective in capturing the long-range dependencies commonly found in physiological signals. By leveraging state-space formulations, S4 models can efficiently represent extended temporal dynamics within PPG sequences. In our MS4 architecture, PPG signals are processed by an S4 backbone that follows the standard S4 configuration, except that we set the maximum sequence length ($l_max$) to 2048 to better capture long-range temporal information. This adjustment enables the model to learn subtle, long-term temporal relationships that are crucial for accurate blood pressure estimation. Similarly, demographic features are fused with the PPG-derived representations at a late stage, following the same fusion strategy used across the other three model variants (ResNet, Inception, LeNet).

\textbf{Loss functions} For all these models, we employ the Mean Squared Error (MSE) loss as our loss function. MSE loss is widely used for regression tasks as it measures the average of the squares of the errors—that is, the average squared difference between the predicted values ($\hat{y}_i$) and the true values ($y_i$). Formally, for a dataset with n samples, the MSE loss is defined as follows:
\begin{equation}
\text{Loss}_{\mathrm{MSE}} = \frac{1}{n} \sum_{i=1}^{n} (y_i - \hat{y}_i)^2
\end{equation}

This loss function penalizes larger errors more heavily, encouraging the model to produce predictions that are as close as possible to the target values.

\section{\uppercase{Experimentation}}

\begin{table*}[htbp]
\centering
\caption{Model Performance Comparison: Calibration-Based vs. Calibration-Free Testing. The values of multimodal models that perform better are highlighted in bold.}

\begin{tabular}{|l|c|c|c|c|c|c|c|}
\hline
\multirow{2}{*}{\textbf{Model}} & \multirow{2}{*}{\textbf{Modality}} 
& \multicolumn{3}{c|}{\textbf{Cal-based (SBP/DBP)}} 
& \multicolumn{3}{c|}{\textbf{Cal-free (SBP/DBP)}} \\
\cline{3-8}
& & \textbf{MAE} & \textbf{std} & \textbf{R²} & \textbf{MAE} & \textbf{std} & \textbf{R²} \\
\hline
ResNet50 & PPG & 5.39/3.31 & 6.98/4.38 & 0.39/0.38 & 6.62/4.32 & 8.03/5.27 & 0.17/0.06 \\
\hline
MResNet50& PPG, Demo & \textbf{5.35/3.24} & \textbf{6.90/4.25} & \textbf{0.40/0.42} & 6.70/\textbf{4.32} & 8.10/\textbf{5.27} & 0.15/\textbf{0.07} \\
\hline
ResNet18 & PPG & 5.57/3.43 & 7.05/4.46 & 0.37/0.36 & 6.60/4.28 & 7.98/5.22 & 0.18/0.09 \\
\hline
MResNet18 & PPG, Demo & \textbf{5.38/3.25} & \textbf{6.97/4.26} & \textbf{0.39/0.41} & \textbf{6.45}/4.32 & \textbf{7.80}/5.29 & \textbf{0.21}/0.07 \\
\hline
Inception & PPG & 6.01/3.77 & 7.41/4.73 & 0.31/0.28 & 6.39/4.31 & 7.84/5.16 & 0.21/0.05 \\
\hline
MInception & PPG, Demo & \textbf{4.75/2.90} & \textbf{6.12/3.84} & \textbf{0.53/0.52} & \textbf{6.34}/4.34 & 7.85/5.34 & \textbf{0.21}/0.00 \\
\hline
S4 & PPG & 6.83/4.57 & 8.18/5.09 & 0.10/0.03 & 6.82/4.53 & 8.14/5.29 & 0.11/0.02 \\
\hline
MS4 & PPG, Demo & 7.10/4.64 & \textbf{7.88/4.89} & 0.07/0.00 & 6.97/4.67 & \textbf{8.06}/5.33 & 0.05/-0.04 \\
\hline
LeNet & PPG & 6.64/4.10 & 8.14/5.10 & 0.17/0.16 & 6.67/4.34 & 8.02/5.28 & 0.17/0.06 \\
\hline
MLeNet & PPG, Demo & 6.76/4.12 & 8.24/5.12 & 0.15/0.15 & \textbf{6.61/4.32} & \textbf{7.98/5.25} & \textbf{0.18/0.09} \\
\hline
\end{tabular}

\label{tab:model_results}
\vspace{-0.1cm} 
\end{table*}

\subsection{Experimental settings}
The neural network models were implemented using the PyTorch deep learning framework and trained on a high-performance computing cluster equipped with four NVIDIA A100 GPUs. We trained models using the Adam optimizer with the following hyperparameters: a batch size of 32, betas of (0.9, 0.999), a weight decay of $1 \times 10^{-8}$, and a fixed learning rate of $2 \times 10^{-5}$. Each model was trained for 100 epochs.

Importantly, our modeling approach is designed so that each neural network is dedicated to predicting a single blood pressure component: systolic blood pressure (SBP) or diastolic blood pressure (DBP). We adopted this one-task-per-model scheme to maintain between SBP and DBP prediction, which reduces potential cross-component interference and simplifying the learning objective for each model. The models were trained independently, following methods established in previous works \citep{pulsedb_paper}, ensuring methodological consistency and understanding of the results.

\subsection{Results}

The evaluation focuses on five representative models: MResNet18, MResNet50, MInception, MLeNet, and MS4. The prefix “M” denotes multimodality, indicating the integration of demographic information (age, sex, and BMI) with PPG signals. Model performance was primarily assessed using mean absolute error (MAE) for both systolic (SBP) and diastolic (DBP) blood pressure, expressed in millimeters of mercury (mmHg). To provide a comprehensive evaluation of accuracy, robustness, and agreement with reference values, additional metrics were employed, including the standard deviation (std) and the coefficient of determination ($R^2$).

For the three best-performing models (ResNet18, ResNet50, Inception), we estimated confidence intervals via 10-fold cross-validation as a substitute for bootstrapping, obtaining confidence intervals for each prediction. Confidence interval plots and waveform-level comparisons (Figures~\ref{fig:top10_modelss} and \ref{fig:top10_modelsd}) were used to visually assess the fidelity of individual segment predictions against ground truth. Furthermore, residual distribution plots (Figure~\ref{fig:residuals_all}) were generated to provide additional insights into model behavior.

As summarized in Table~\ref{tab:model_results}, all models were tested under both calibration-based and calibration-free settings. Results consistently demonstrate that the integration of demographic data enhances prediction performance, particularly in calibration-free conditions where generalizability is more challenging. Among the tested architectures, MInception achieved the best calibration-based results, with an MAE of \textbf{4.75/2.90} mmHg for SBP/DBP, standard deviations of \textbf{6.12/3.84}, and $R^2$ values of \textbf{0.53/0.52}. For calibration-free evaluation, MInception reached an MAE of \textbf{6.34/4.34}, while MResNet18 achieved \textbf{6.45/4.32}, both demonstrating robustness superior to unimodal baselines such as ResNet18 (\textbf{6.60/4.28}) and ResNet50 (\textbf{6.62/4.32}). Overall, these findings highlight the critical role of demographic information in achieving reliable cuffless blood pressure estimation and underscore the importance of evaluating models across calibration scenarios.

\section{Discussion}
The neural network architecture and the inclusion of demographic information play important roles in the accuracy of blood pressure prediction from PPG signals. Models designed to extract multi-scale features and fuse heterogeneous data sources (such as Inception-based architectures) benefit most from the integration of demographic data. In contrast, models focusing solely on sequential patterns or with simpler structures such as S4 are less able to utilize the added value of demographic features. We detail the performance and characteristics of each network setting:
\begin{figure}[htbp]
    \centering
    \begin{minipage}{0.48\textwidth}
        \centering
        \includegraphics[width=\linewidth]{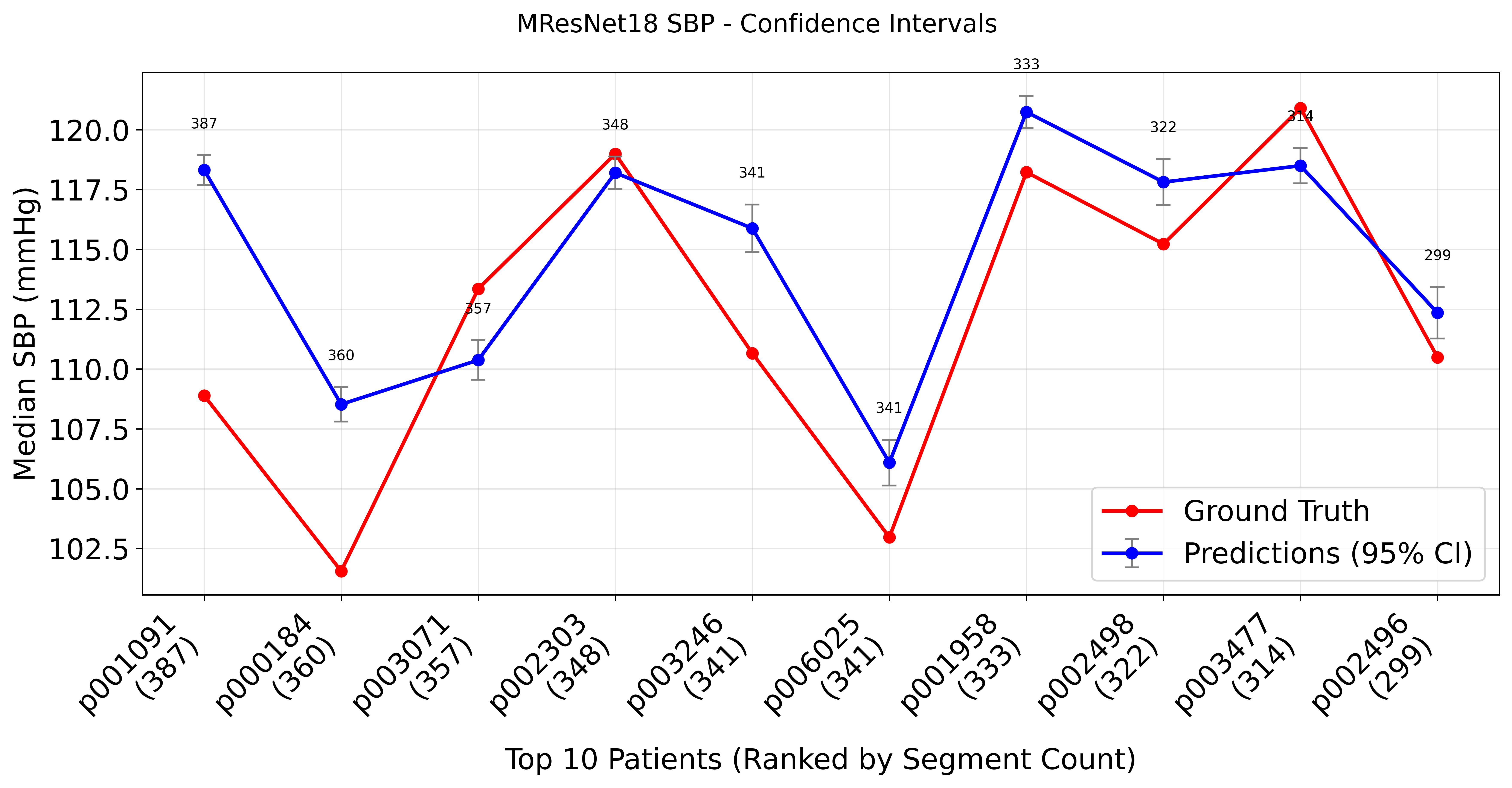}
    \end{minipage}\hfill

    \vspace{0.1cm} 
    
    \begin{minipage}{0.48\textwidth}
        \centering
        \includegraphics[width=\linewidth]{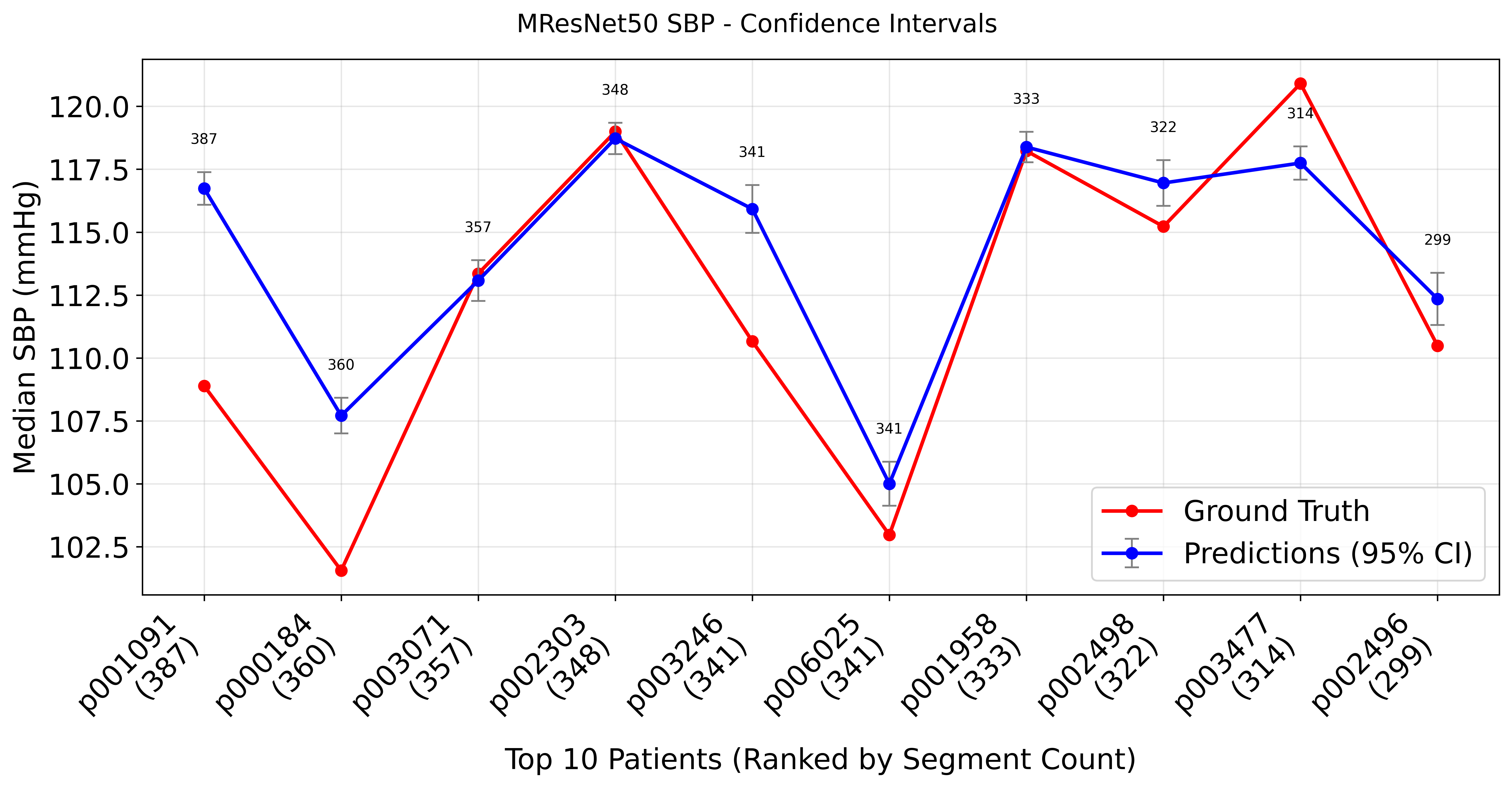}
    \end{minipage}

    \vspace{0.1cm} 

    \begin{minipage}{0.48\textwidth}
        \centering
        \includegraphics[width=\linewidth]{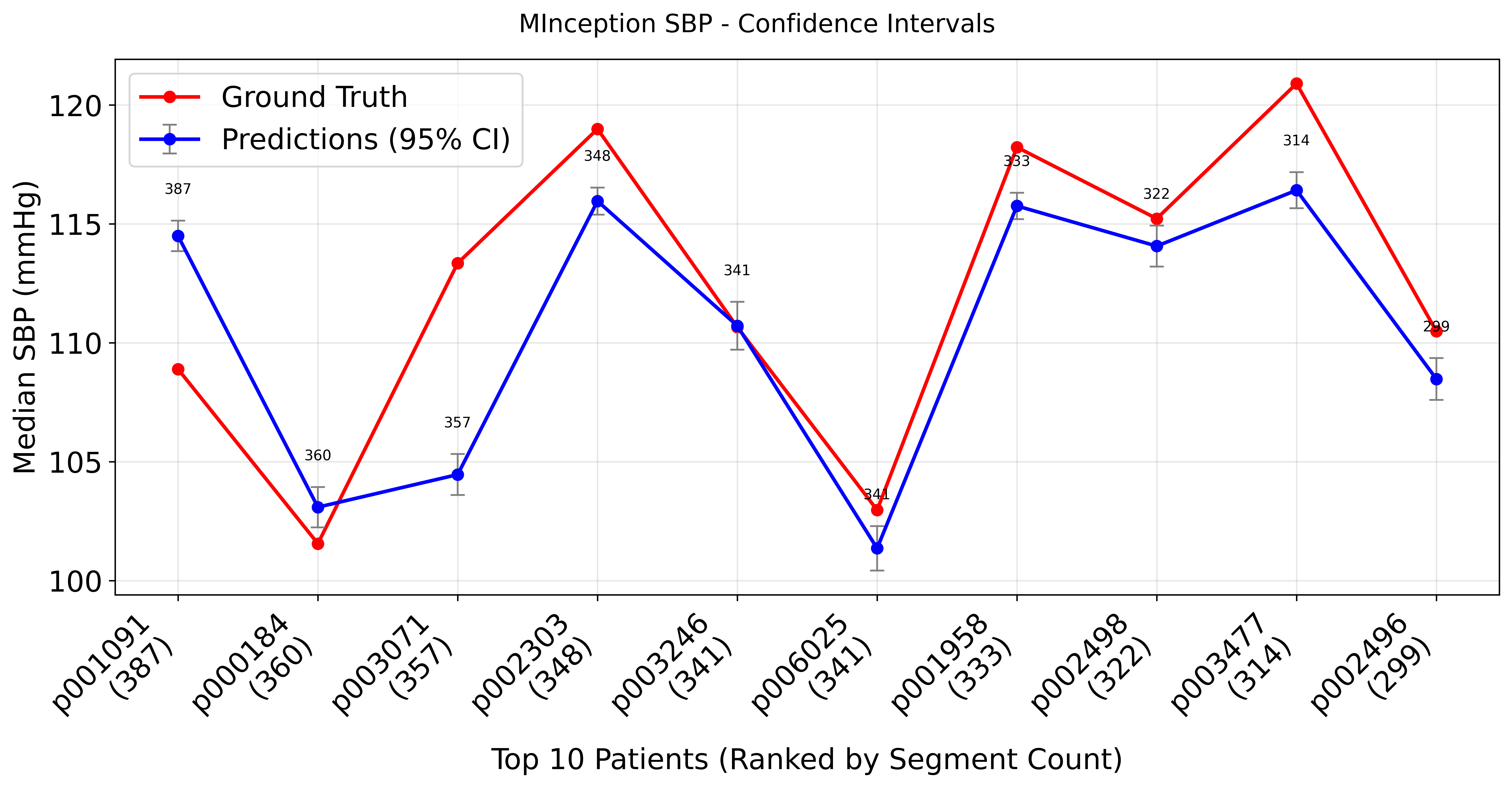}
    \end{minipage}
    
    \vspace{0.1cm} 
    
    \caption{Prediction with confidence intervals vs. ground truth of SBP for the top 10 patients with the most segments using: MResNet18-1D, MResNet50-1D, MInception-1D.}
    \label{fig:top10_modelss}

\vspace{0.1cm} 
\end{figure}

\begin{figure}[htbp]
    \centering
    \begin{minipage}{0.48\textwidth}
        \centering
        \includegraphics[width=\linewidth]{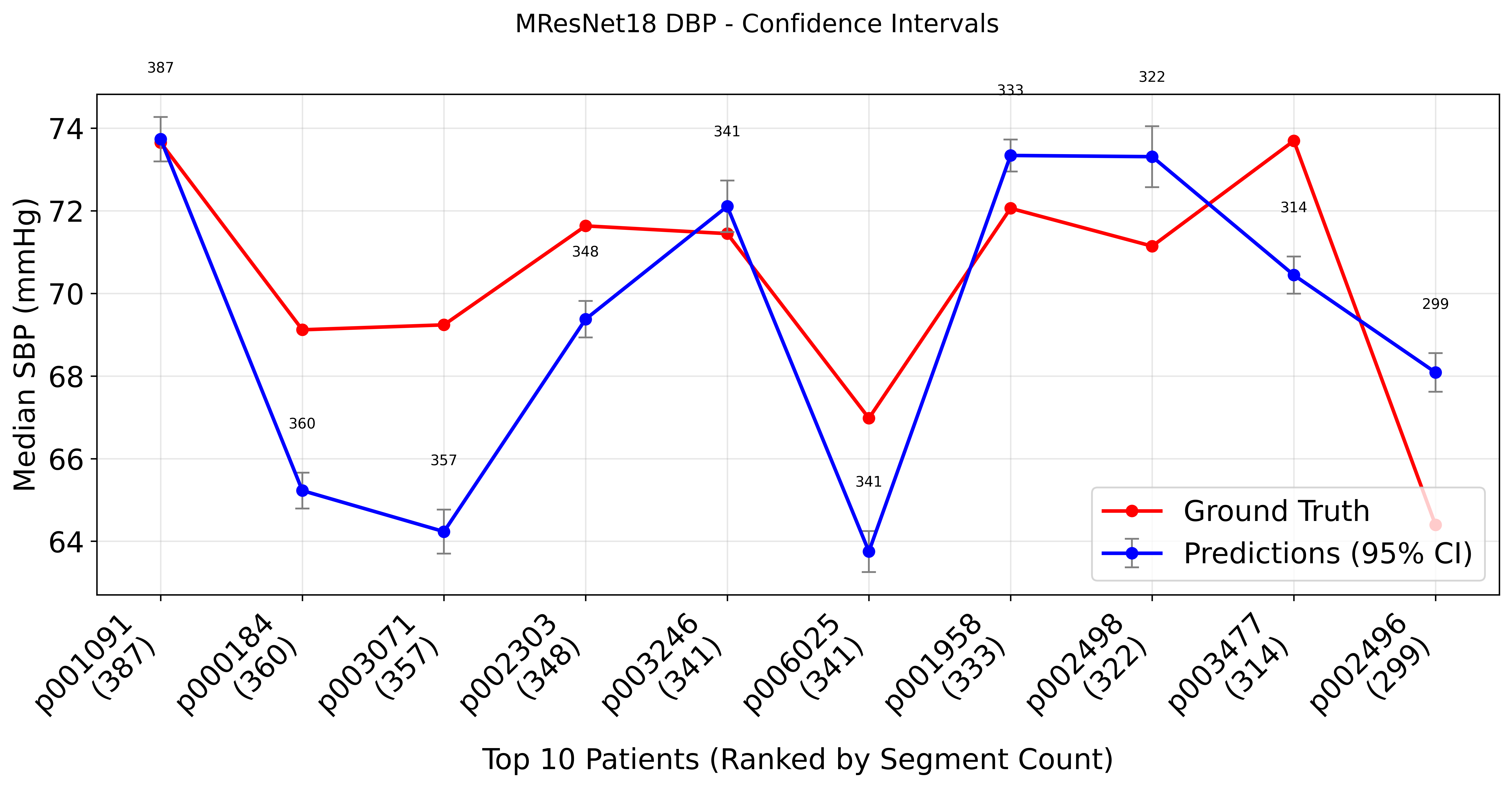}
    \end{minipage}\hfill

    \vspace{0.1cm} 
    
    \begin{minipage}{0.48\textwidth}
        \centering
        \includegraphics[width=\linewidth]{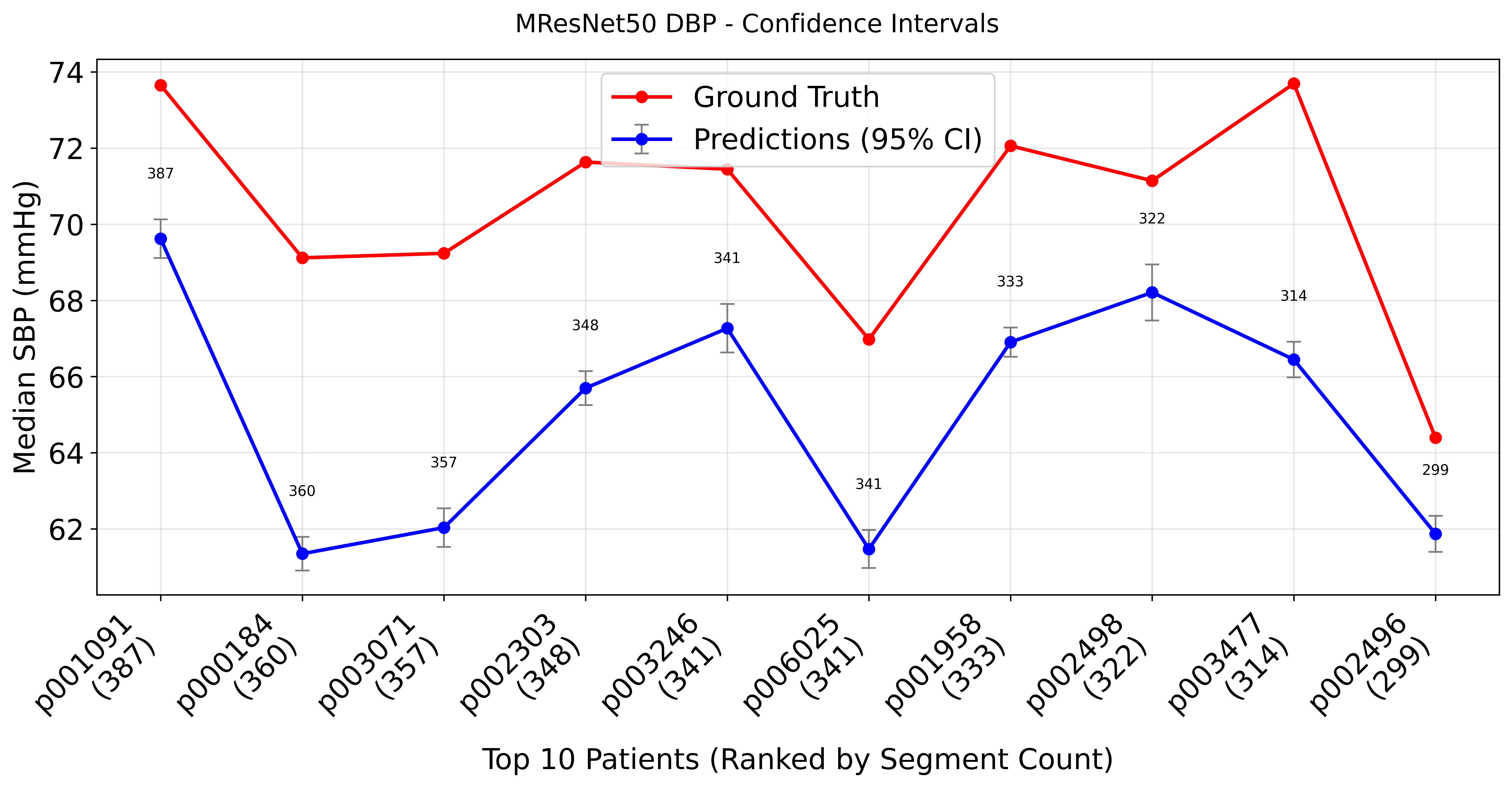}
    \end{minipage}

    \vspace{0.1cm} 

    \begin{minipage}{0.48\textwidth}
        \centering
        \includegraphics[width=\linewidth]{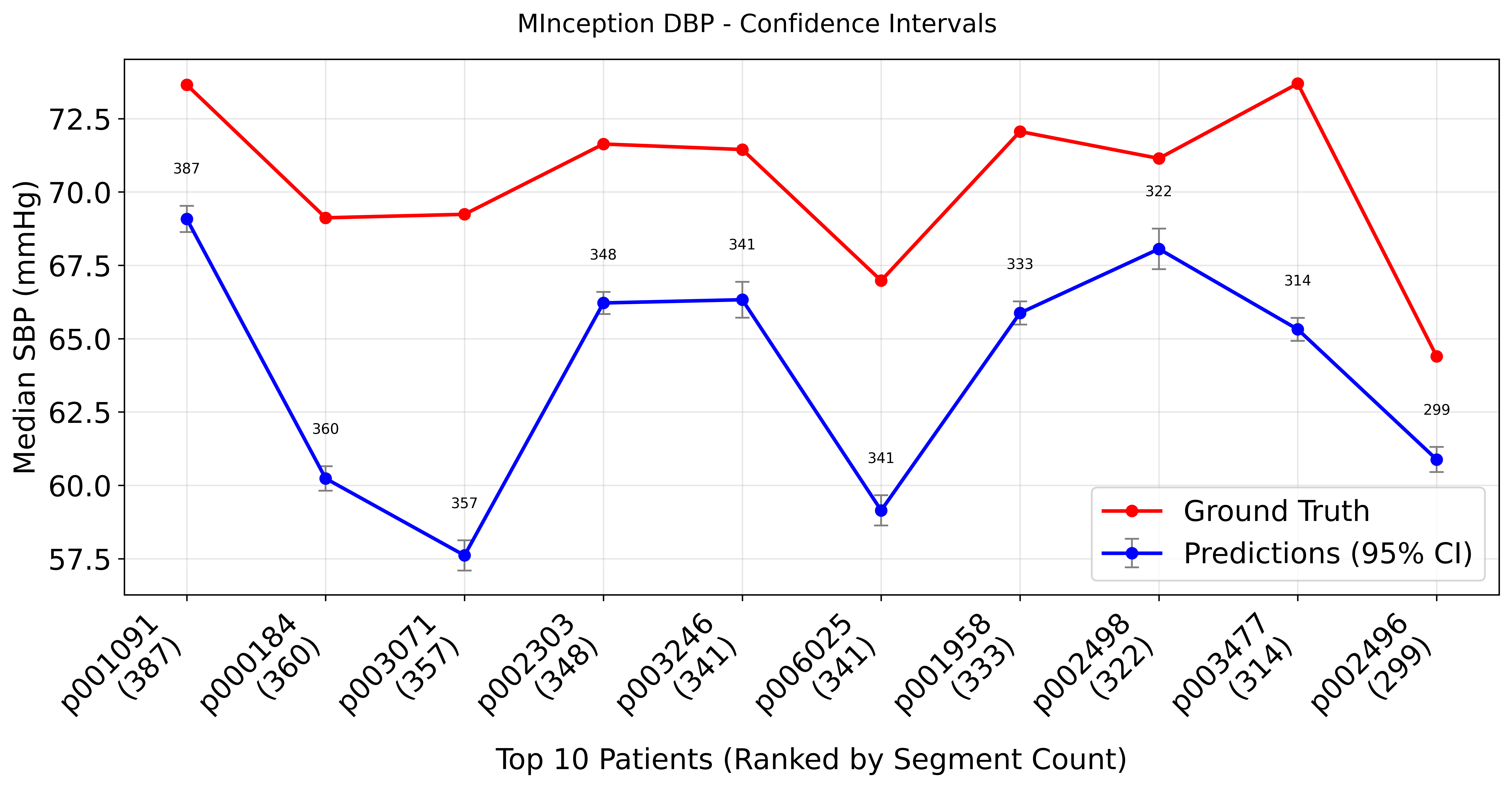}
    \end{minipage}

    \caption{Prediction with confidence intervals vs. ground truth of DBP for the top 10 patients with the most segments using: MResNet18-1D, MResNet50-1D, MInception-1D.}
    \label{fig:top10_modelsd}

\vspace{0.1cm} 
\end{figure}

\textbf{PPG-only Settings:}
In PPG-only settings, none of these models meet the required standards. The calibration-based ResNet50-1D model achieves the best performance among all tested architectures. This is because ResNet50-1D is the most complex model in this group, allowing it to learn more features from the raw PPG signal. The residual connections inherent in ResNet architectures also help mitigate issues such as vanishing or exploding gradients, thus stabilizing and enhancing the learning process, especially as the network depth increases. Consequently, ResNet50 achieves MAE of 5.39/3.31 mmHg (SBP/DBP).  In the calibration-free setting, the Inception-1D model demonstrates superior performance with an MAE of 6.01/3.77 mmHg. This advantage stems from Inception's unique capability of extracting multi-scale and multi-dimensional features through parallel convolutional operations of varying kernel sizes. This flexibility enables Inception to adapt more effectively to unseen data. It can capture features at different levels and is less likely to overfit to validation data.

\begin{figure*}[!htbp]
    \centering
    \begin{minipage}{0.45\textwidth}
        \centering
        \includegraphics[width=\linewidth]{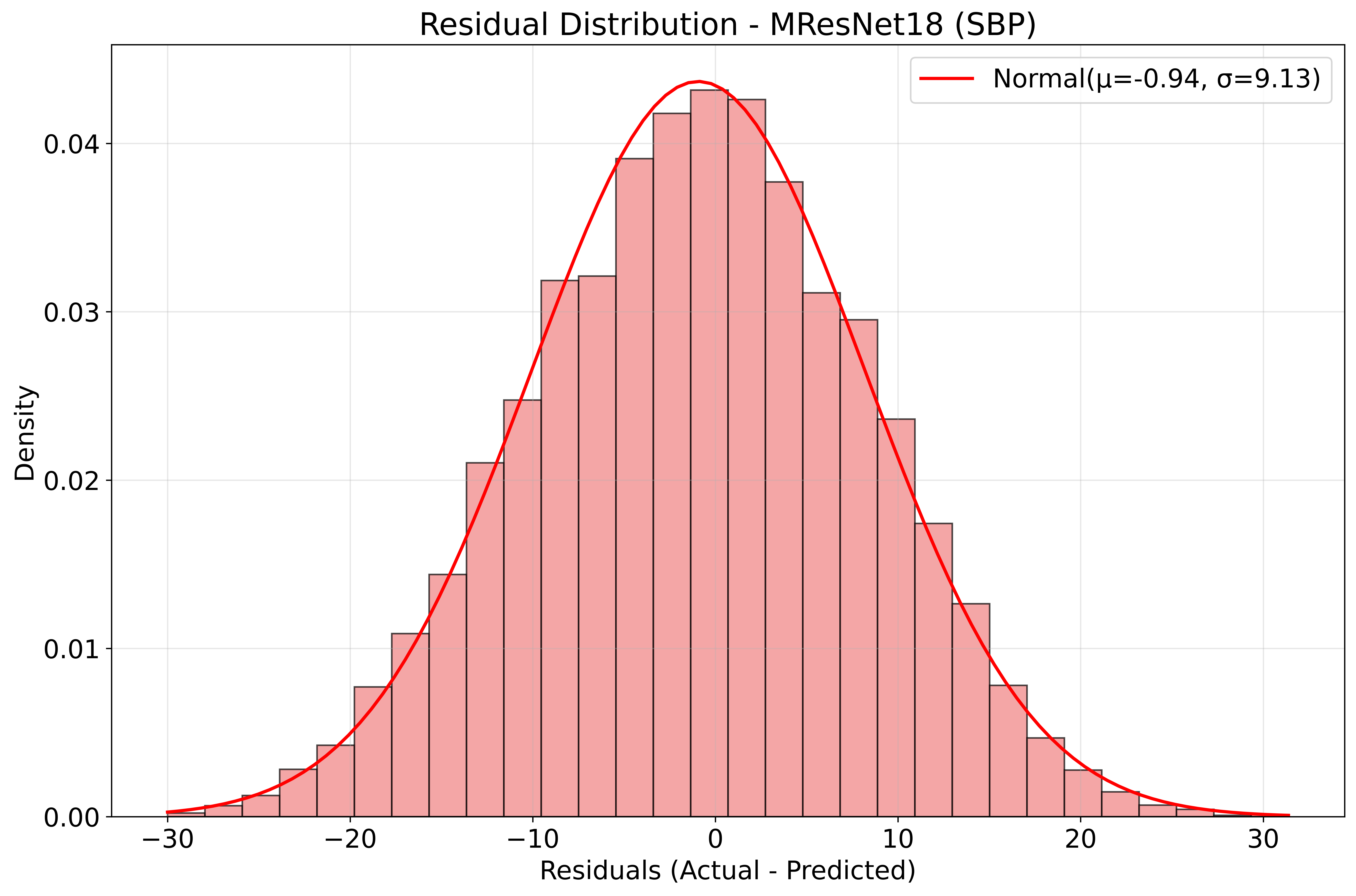}
    \end{minipage}\hfill
    \begin{minipage}{0.45\textwidth}
        \centering
        \includegraphics[width=\linewidth]{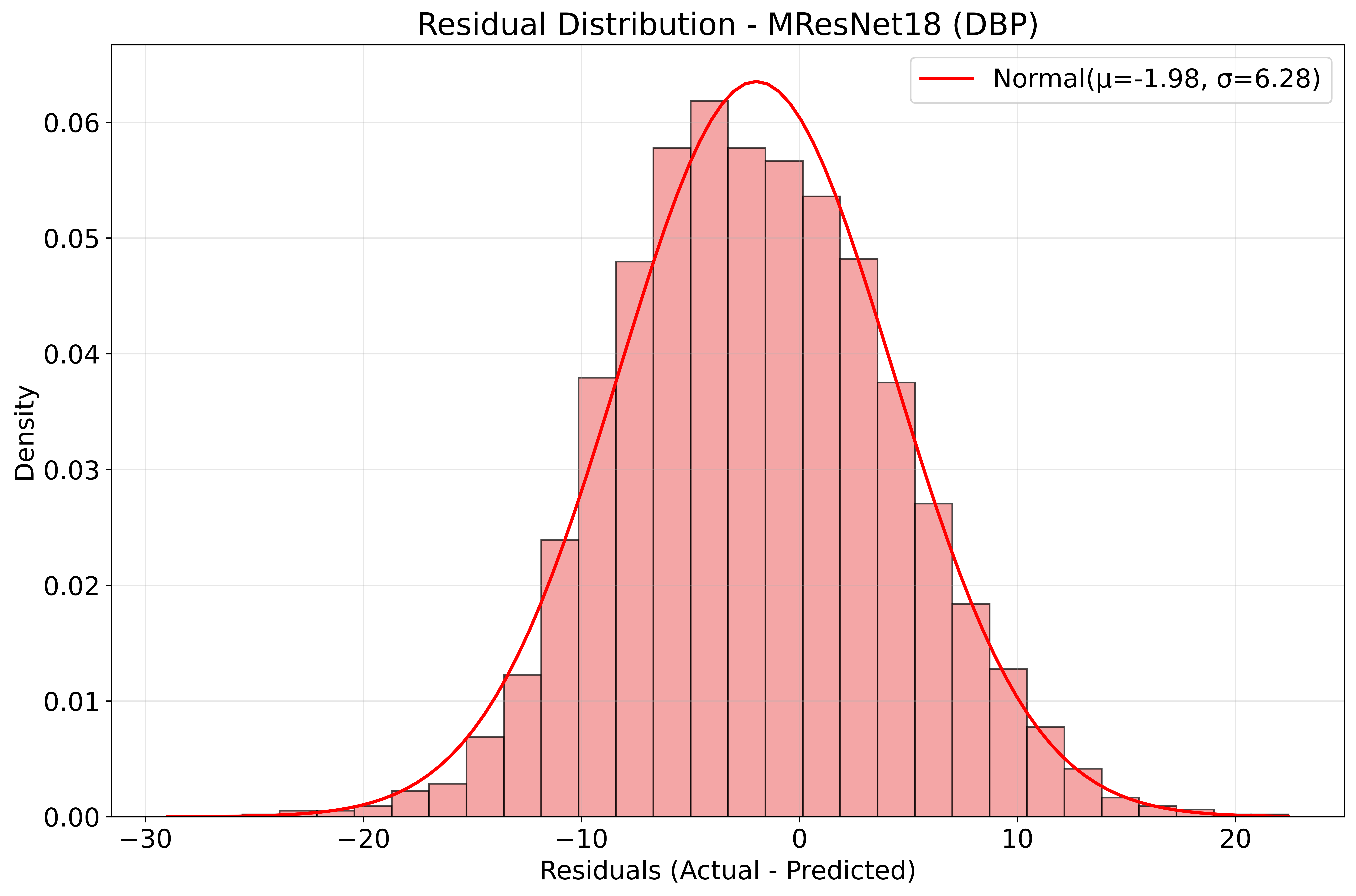}
    \end{minipage}


    \begin{minipage}{0.45\textwidth}
        \centering
        \includegraphics[width=\linewidth]{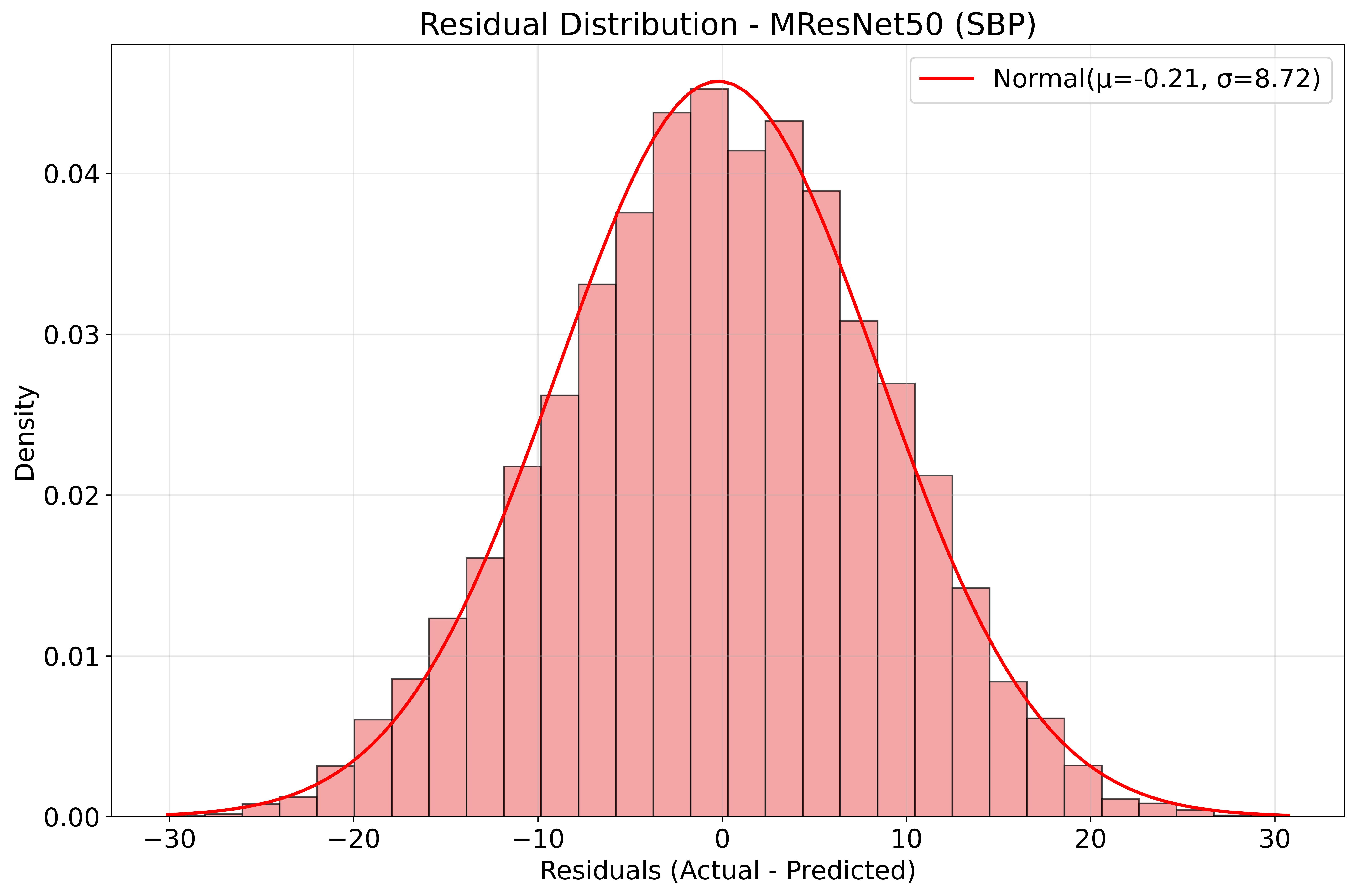}
    \end{minipage}\hfill
    \begin{minipage}{0.45\textwidth}
        \centering
        \includegraphics[width=\linewidth]{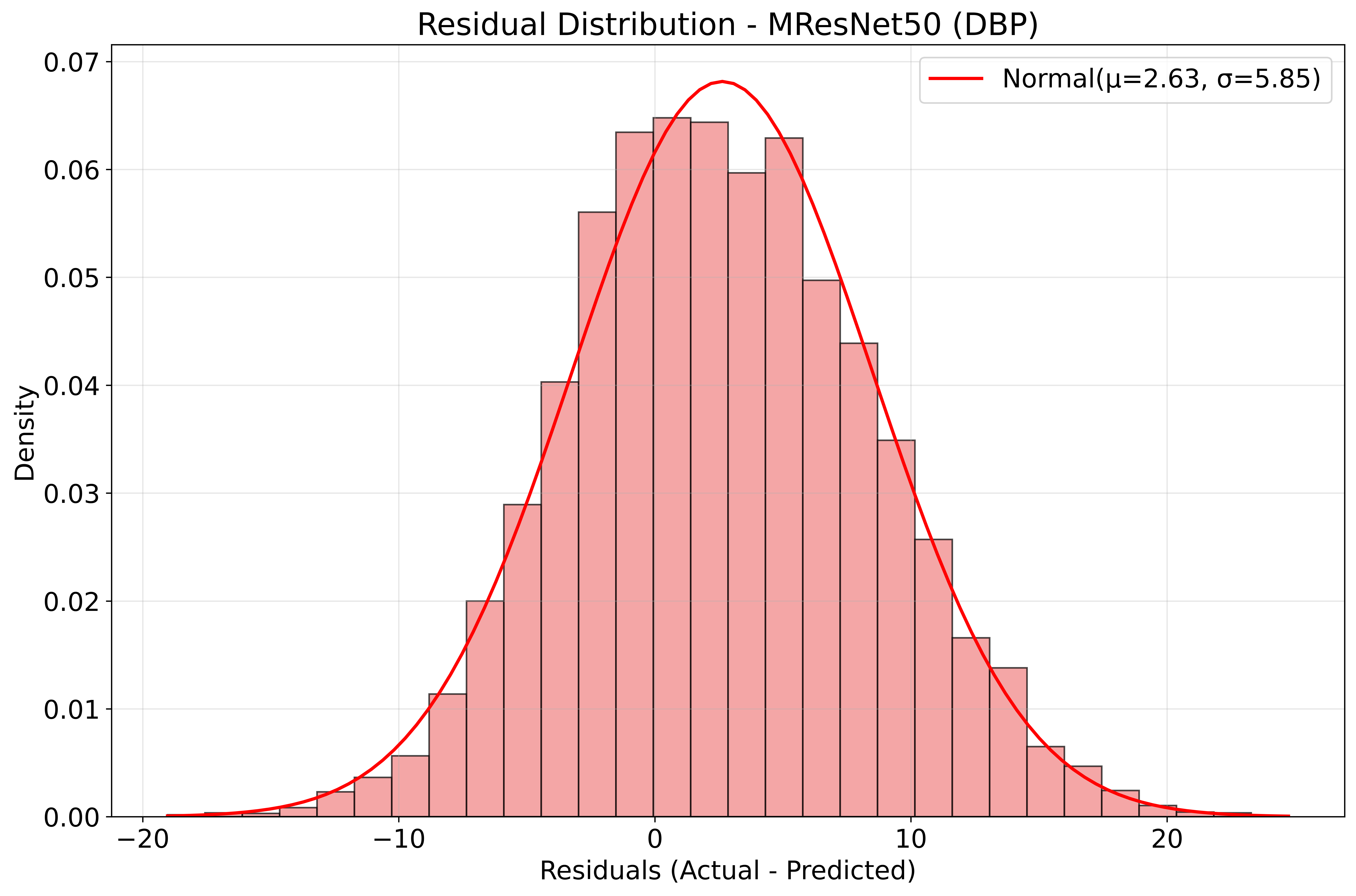}
    \end{minipage}


    \begin{minipage}{0.45\textwidth}
        \centering
        \includegraphics[width=\linewidth]{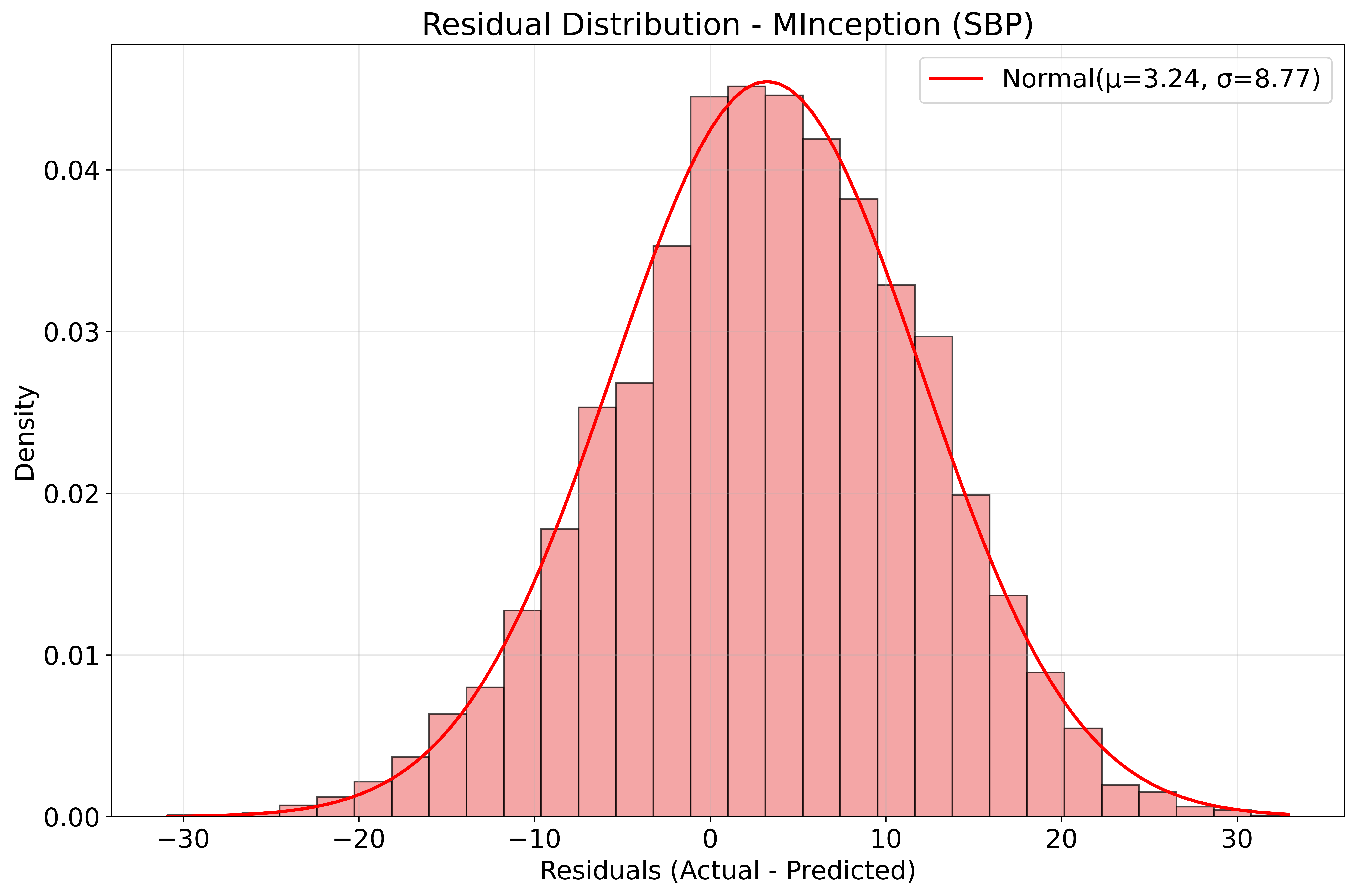}
    \end{minipage}\hfill
    \begin{minipage}{0.45\textwidth}
        \centering
        \includegraphics[width=\linewidth]{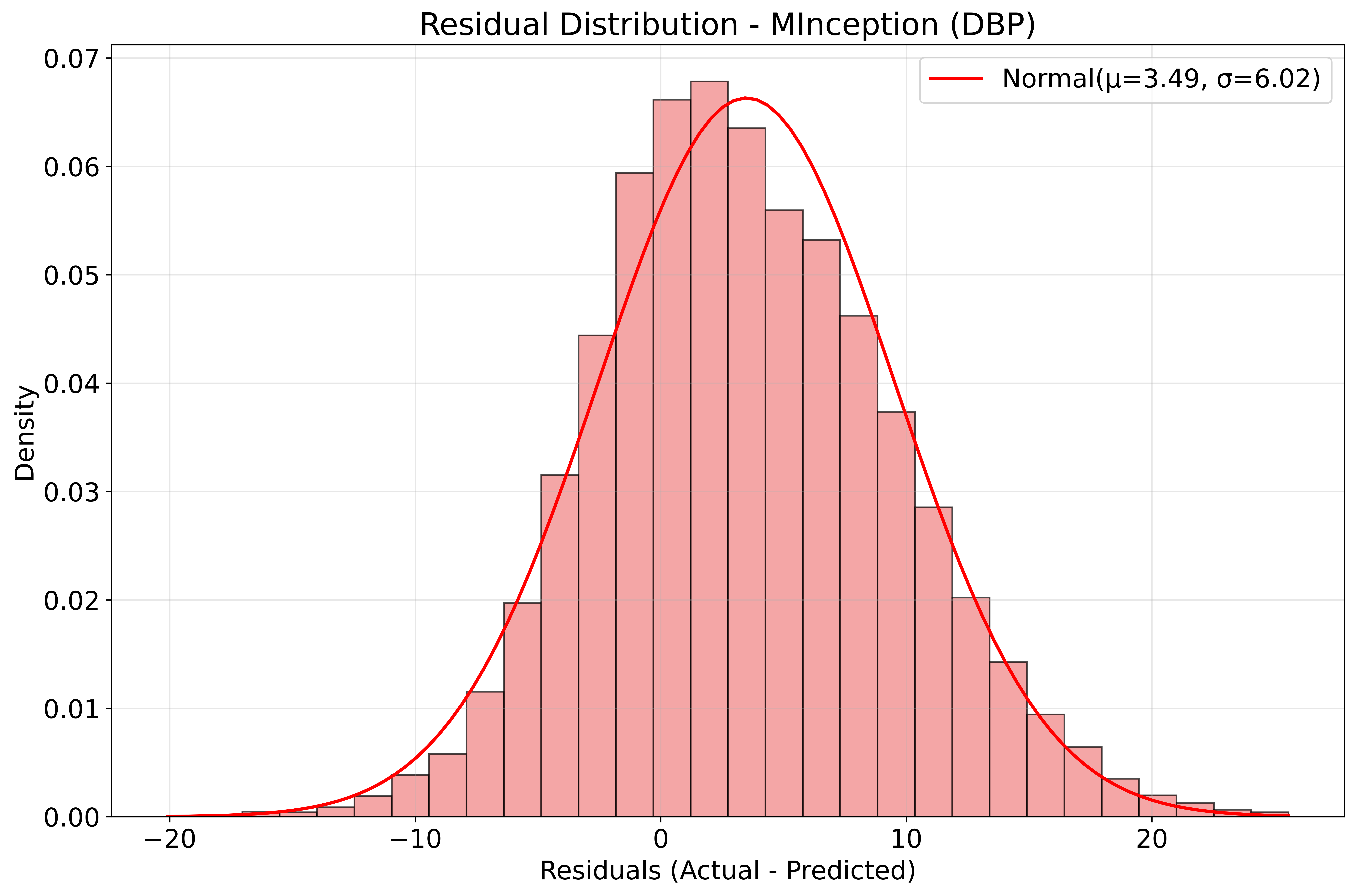}
    \end{minipage}
    \caption{Residual distributions of MResNet18-1D, MResNet50-1D, and MInception-1D for SBP (left) and DBP (right).}
    \label{fig:residuals_all}

\vspace{0.1cm} 
    
\end{figure*}

\textbf{PPG \& Demographic Settings:}
When using both PPG signals and demographic features, MInception achieves MAE of 4.75/2.90 mmHg, meeting accuracy comparable to AAMI/ISO limits (mean error \textless5 mmHg and standard deviation \textless8 mmHg), yielding an MAE of 4.75/2.90 mmHg (SBP/DBP), std of 6.12/3.84 mmHg (SBP/DBP) in calibration-based and 6.34/4.34 mmHg in calibration-free scenarios. The Inception architecture demonstrated substantial performance gains from demographic integration, with 21\% MAE reduction. This is particularly beneficial when these features are concatenated with demographic information. While ResNet-based models excel in extracting features from a single modality, their performance may be limited when integrating over a wider input feature range. This is because single-modality feature extraction may introduce bias or fail to exploit the connection between PPG and demographic features.

\textbf{Comparison: PPG vs. PPG + Demographic:}

For ResNet-based models, adding demographic variables improves performance in both settings by an average of 3\% in the calibration-based experiments and 1\% in the calibration-free experiments. Gains in both settings suggest that demographic features contribute patient-specific priors that help the model learn more robust relationships between PPG and blood pressure, improving generalization to subjects especially in calibration-based conditions.

Inception-based models derive the greatest benefit from the inclusion of demographic data, 23\% improvement in the calibration-based. Their architectural design, which supports the extraction of multi-scale and multi-dimensional features, enables effective fusion of complex demographic information. This results in a clearer representation of patient data, further boosting predictive performance.

It was expected that a simple LeNet would be unable to distinguish demographic information, however, clear improvements were shown when demographic features are incorporated via late fusion: with a 7\% improvement in the calibration-free setting. These improvements likely arise because demographic variables provide informative global priors that a low-capacity CNN can utilize to reduce hypothesis space and stabilize optimization. In this configuration, demographic features act as subject-level bias or conditioning vectors that complements LeNet’s local morphology-focused features, helping to form unique PPG patterns that are difficult to separate using signal alone.

Interestingly, the S4 model is the only architecture that does not benefit from the inclusion of demographic data, showing an average performance drop of approximately 3\%. Although a late fusion strategy avoids the mixture of modalities which may cause the model to struggle. There may still be interplay between feature dominance, data characteristics, and optimization dynamics.
The S4 backbone has a large 512-dimensional output, which produces rich temporal representations that dominate the learning process. When demographic features are fused via simple concatenation, their contribution becomes marginal and the fusion layers tend to prioritize the stronger temporal features while treating static information as noise. Furthermore, demographic variables such as age, sex, or BMI often exhibit weak correlations or dataset-specific correlations with physiological signals like PPG. As a result, the inclusion of such features may slightly destabilize the learned decision boundary or introduce mild overfitting.
Additionally, the late fusion mechanism used here assumes that temporal and static features reside in a compatible feature space and can be linearly combined. However, these modalities are inherently heterogeneous: one dynamic and high-dimensional, the other static and low-dimensional, making simple concatenation insufficient for capturing their conditional dependencies when combined with increased parameterization of the fusion module and potential optimization imbalance between branches, the model's generalization can slightly degrade.

\textbf{Residual and Confidence Interval Analysis:}
To further examine model behavior, residual distributions (ground truth minus prediction) and confidence intervals were analyzed for three representative architectures: MResNet18-1D, MResNet50-1D, and MInception-1D. Figure~\ref{fig:residuals_all} illustrates the residual distributions for SBP and DBP predictions. All three models exhibit approximately normal distributions centered near zero, indicating unbiased estimates. Among them, MInception-1D shows the narrowest spread of residuals, consistent with its MAE and $R^2$ values, while MResNet18-1D demonstrates relatively stable but slightly broader distributions compared to MResNet50-1D and MInception-1D.

Figure~\ref{fig:top10_modelss} and Figure~\ref{fig:top10_modelsd} presents the prediction with confidence interval versus ground truth for the top 10 patients with the largest number of segments. The results show that predictions closely track ground truth across patients, with the 95\% confidence intervals overlapping the reference values in DBP. These analyses highlight that incorporating demographic features not only improves point-estimate accuracy but also reduces uncertainty in patient-level predictions, reinforcing the potential for multimodality approaches.

\textbf{Clinical Relevance and Intended Use:}
The MInception architecture exhibits commendable performance with low error margins , yielding MAE of 4.75/2.90 mmHg (SBP/DBP), std of 6.12/3.84 mmHg (SBP/DBP) in calibration-based setting which is accuracy comparable to AAMI/ISO 81060-2 error limits of blood pressure monitoring within the studied population. These results indicate that the proposed model achieves accuracy levels suitable for home-based blood pressure assessment.

In practical applications, MInception can be integrated into wearable or cuffless monitoring systems, providing continuous, non-invasive blood pressure estimation. Such capability is particularly valuable for patients requiring long-term monitoring, as well as for preventive healthcare where early detection of abnormal blood pressure fluctuations is critical.

\section{\uppercase{Limitations and Future Work}}
We provide a benchmark of multiple neural networks for cuffless BP estimation from PPG with demographic data. Our benchmark uses a healthy-adult dataset (NBPDB).

\textbf{Limitations:} The modest calibration-free gains ($<3\%$ for most models) contrast with the larger calibration-based improvements ($3-23\%$). Two factors may explain this discrepancy. First, the highly imbalanced data structure: eighty thousand segments from only hundreds of subjects, likely causes overfitting to subject-specific demographic patterns, thereby limiting discriminative power across cal-free individuals. Second, the architecture-dependent results (e.g., a 23\% gain for Inception v.s. a $-3\%$ change for S4) suggest that simple late-fusion strategies may be poorly suited for sequential models such as S4. Finally, several networks still exhibit low R² values, indicating that substantial work remains to improve both predictive accuracy and generalization.

\textbf{Future work:} (i) Expand the corpus to include abnormal signals and external datasets to assess robustness and clinical transferability. (ii) Systematically study demography-PPG fusion beyond late fusion—e.g., attention/gating, cross-modal conditioning, and uncertainty-aware fusion. (iii) Explore recent state-of-the-art (SOTA) sequence models, including transformer-based and long-context architectures, for improved temporal modeling and multimodal integration, under both calibration-based and calibration-free protocols.

\section{\uppercase{Conclusion}}

We present a standardized healthy-adult blood pressure benchmarking subset (NBPDB) derived from MIMIC-III and VitalDB, along with a 
benchmarking framework for cuffless blood pressure estimation from PPG in healthy adults, evaluating models under calibration-based and calibration-free settings. Across all architectures, PPG-only models did not meet AAMI/ISO accuracy, whereas demography-aware models that integrate age, sex, and BMI consistently improved performance and robustness. 
Under the calibration-based protocol, our best demography-aware model (MInception) meets AAMI/ISO criteria, achieving an MAE of 4.75/2.90 mmHg (SBP/DBP), supporting the clinical value of incorporating physiologically grounded demographic factors. Meeting this clinical-grade standard indicates that such models could be translated into wearable blood pressure monitoring systems capable of providing continuous, non-invasive, and standards-compliant BP estimation.



\bibliographystyle{apalike}
{\small
\bibliography{example}}

@article{mehta2024examining,
  author = {Mehta, S. and Kwatra, N. and Jain, M. and McDuff, D.},
  title = {Examining the challenges of blood pressure estimation via photoplethysmogram},
  journal = {Scientific Reports},
  volume = {14},
  number = {1},
  pages = {18318},
  year = {2024},
  doi = {10.1038/s41598-024-68862-1}
}

@article{pat_paper,
author = {Zhang, Guanqun and Gao, Mingwu and Xu, Da and Olivier, N. Bari and Mukkamala, Ramakrishna},
title = {Pulse arrival time is not an adequate surrogate for pulse transit time as a marker of blood pressure},
journal = {Journal of Applied Physiology},
volume = {111},
number = {6},
pages = {1681-1686},
year = {2011},
doi = {10.1152/japplphysiol.00980.2011},
note ={PMID: 21960657},
URL = { 
        https://doi.org/10.1152/japplphysiol.00980.2011
},
eprint = { 
        https://doi.org/10.1152/japplphysiol.00980.2011
}
}

@article{HU20239,
title = {Validating cuffless continuous blood pressure monitoring devices},
journal = {Cardiovascular Digital Health Journal},
volume = {4},
number = {1},
pages = {9-20},
year = {2023},
issn = {2666-6936},
doi = {https://doi.org/10.1016/j.cvdhj.2023.01.001},
url = {https://www.sciencedirect.com/science/article/pii/S2666693623000014},
author = {Jiun-Ruey Hu and Gabrielle Martin and Sanjna Iyengar and Lara C. Kovell and Timothy B. Plante and Noud van Helmond and Richard A. Dart and Tammy M. Brady and Ruth-Alma N. Turkson-Ocran and Stephen P. Juraschek}
}

@misc{raza2025neuromoetransformerbasedmixtureofexpertsframework,
      title={NeuroMoE: A Transformer-Based Mixture-of-Experts Framework for Multi-Modal Neurological Disorder Classification}, 
      author={Wajih Hassan Raza and Aamir Bader Shah and Yu Wen and Yidan Shen and Juan Diego Martinez Lemus and Mya Caryn Schiess and Timothy Michael Ellmore and Renjie Hu and Xin Fu},
      year={2025},
      eprint={2506.14970},
      archivePrefix={arXiv},
      primaryClass={eess.IV},
      url={https://arxiv.org/abs/2506.14970}, 
}

@article{pwv_paper,
AUTHOR = {Zhou, Zi-Bo and Cui, Tian-Rui and Li, Ding and Jian, Jin-Ming and Li, Zhen and Ji, Shou-Rui and Li, Xin and Xu, Jian-Dong and Liu, Hou-Fang and Yang, Yi and Ren, Tian-Ling},
TITLE = {Wearable Continuous Blood Pressure Monitoring Devices Based on Pulse Wave Transit Time and Pulse Arrival Time: A Review},
JOURNAL = {Materials},
VOLUME = {16},
YEAR = {2023},
NUMBER = {6},
ARTICLE-NUMBER = {2133},
URL = {https://www.mdpi.com/1996-1944/16/6/2133},
PubMedID = {36984013},
ISSN = {1996-1944},
DOI = {10.3390/ma16062133}
}

@article{yusheng,
title = {Interference source-based quality assessment method for postauricular photoplethysmography signals},
journal = {Biomedical Signal Processing and Control},
volume = {84},
pages = {104751},
year = {2023},
issn = {1746-8094},
doi = {https://doi.org/10.1016/j.bspc.2023.104751},
url = {https://www.sciencedirect.com/science/article/pii/S1746809423001842},
author = {Yusheng Qi and Aihua Zhang and Yurun Ma and Huidong Wang and Jiaqi Li}
}

@misc{nie2024reviewdeeplearningmethods,
      title={A Review of Deep Learning Methods for Photoplethysmography Data}, 
      author={Guangkun Nie and Jiabao Zhu and Gongzheng Tang and Deyun Zhang and Shijia Geng and Qinghao Zhao and Shenda Hong},
      year={2024},
      eprint={2401.12783},
      archivePrefix={arXiv},
      primaryClass={cs.AI},
      url={https://arxiv.org/abs/2401.12783}, 
}

@article{tale2021sphygmomanometers,
    author = {Tale, Shreya and Joshi, Swanand and Ambad, Ranjit Sidram and Bankar, Nandkishor},
    title = {Sphygmomanometers: Technological Advancements and Significance in Diagnostics},
    journal = {Natural Volatiles \& Essential Oils},
    year = {2021},
    volume = {8},
    number = {5},
    pages = {1453--1456}
}

@article{huang2024validation,
  author    = {Huang, Y. and He, Y. and Song, Z. and Gao, K. and Zheng, Y.},
  title     = {Validation of deep learning models for cuffless blood pressure estimation on a large benchmarking dataset},
  journal   = {Connected Health and Telemedicine},
  year      = {2024},
  volume    = {3},
  pages     = {300002},
  doi       = {10.20517/chatmed.2023.23},
  url       = {http://dx.doi.org/10.20517/chatmed.2023.23}
}

@article{elgendi2019photoplethysmography,
  author = {Elgendi, Mohamed and Fletcher, Richard and Liang, Yongbo and Howard, Newton and Lovell, Nigel H. and Abbott, Derek and Lim, Kenneth and Ward, Rabab},
  title = {The use of photoplethysmography for assessing hypertension},
  journal = {npj Digital Medicine},
  volume = {2},
  pages = {60},
  year = {2019},
  doi = {10.1038/s41746-019-0136-7}
}

@article{cabanas2022skin,
  author = {Cabanas, A. M. and Fuentes-Guajardo, M. and Latorre, K. and León, D. and Martín-Escudero, P.},
  title = {Skin Pigmentation Influence on Pulse Oximetry Accuracy: A Systematic Review and Bibliometric Analysis},
  journal = {Sensors},
  volume = {22},
  number = {9},
  pages = {3402},
  year = {2022},
  doi = {10.3390/s22093402},
  url = {https://doi.org/10.3390/s22093402}
}

@article{yan2005reduction,
  author = {Yan, Y. S. and Poon, C. C. and Zhang, Y. T.},
  title = {Reduction of motion artifact in pulse oximetry by smoothed pseudo Wigner-Ville distribution},
  journal = {Journal of NeuroEngineering and Rehabilitation},
  volume = {2},
  pages = {3},
  year = {2005},
  doi = {10.1186/1743-0003-2-3},
  url = {https://doi.org/10.1186/1743-0003-2-3}
}

@article{chen2024ppg,
  author = {Chen, G and Zou, L and Ji, Z},
  title = {A review: Blood pressure monitoring based on PPG and circadian rhythm},
  journal = {APL Bioengineering},
  volume = {8},
  number = {3},
  pages = {031501},
  year = {2024},
  doi = {10.1063/5.0206980}
}

@article{chu2023noninvasive,
  author = {Chu, Y. and Tang, K. and Hsu, Y. C. and Huang, T. and Wang, D. and Li, W. and Savitz, S. I. and Jiang, X. and Shams, S.},
  title = {Non-invasive arterial blood pressure measurement and SpO2 estimation using PPG signal: a deep learning framework},
  journal = {BMC Medical Informatics and Decision Making},
  volume = {23},
  number = {1},
  pages = {131},
  year = {2023},
  doi = {10.1186/s12911-023-02215-2}
}

@ARTICLE{escobarpaper,
AUTHOR={Escobar-Restrepo, Braiam  and Torres-Villa, Robinson  and Kyriacou, Panayiotis A. },     
TITLE={Evaluation of the Linear Relationship Between Pulse Arrival Time and Blood Pressure in ICU Patients: Potential and Limitations},
JOURNAL={Frontiers in Physiology},
VOLUME={Volume 9 - 2018},
YEAR={2018},
URL={https://www.frontiersin.org/journals/physiology/articles/10.3389/fphys.2018.01848},
DOI={10.3389/fphys.2018.01848},
ISSN={1664-042X}
}

@misc{gu2022efficientlymodelinglongsequences,
      title={Efficiently Modeling Long Sequences with Structured State Spaces}, 
      author={Albert Gu and Karan Goel and Christopher Ré},
      year={2022},
      eprint={2111.00396},
      archivePrefix={arXiv},
      primaryClass={cs.LG},
      url={https://arxiv.org/abs/2111.00396}, 
}

@Article{sadrawi,
AUTHOR = {Sadrawi, Muammar and Lin, Yin-Tsong and Lin, Chien-Hung and Mathunjwa, Bhekumuzi and Fan, Shou-Zen and Abbod, Maysam F. and Shieh, Jiann-Shing},
TITLE = {Genetic Deep Convolutional Autoencoder Applied for Generative Continuous Arterial Blood Pressure via Photoplethysmography},
JOURNAL = {Sensors},
VOLUME = {20},
YEAR = {2020},
NUMBER = {14},
ARTICLE-NUMBER = {3829},
URL = {https://www.mdpi.com/1424-8220/20/14/3829},
PubMedID = {32660088},
ISSN = {1424-8220},
DOI = {10.3390/s20143829}
}

@misc{moulaeifard2025generalizabledeeplearningphotoplethysmographybased,
      title={Generalizable deep learning for photoplethysmography-based blood pressure estimation -- A Benchmarking Study}, 
      author={Mohammad Moulaeifard and Peter H. Charlton and Nils Strodthoff},
      year={2025},
      eprint={2502.19167},
      archivePrefix={arXiv},
      primaryClass={cs.LG},
      url={https://arxiv.org/abs/2502.19167}, 
}

@article{ELHAJJ2021102301,
title = {Deep learning models for cuffless blood pressure monitoring from PPG signals using attention mechanism},
journal = {Biomedical Signal Processing and Control},
volume = {65},
pages = {102301},
year = {2021},
issn = {1746-8094},
doi = {https://doi.org/10.1016/j.bspc.2020.102301},
url = {https://www.sciencedirect.com/science/article/pii/S1746809420304201},
author = {C El-Hajj and PA Kyriacou}
}

@article{flint2019effect,
  author = {Flint, A. C. and Conell, C. and Ren, X. and Banki, N. M. and Chan, S. L. and Rao, V. A. and Zarefar, A. and Erani, D. M. and Nguyen-Huynh, M. N. and Sidney, S.},
  title = {Effect of systolic and diastolic blood pressure on cardiovascular outcomes},
  journal = {New England Journal of Medicine},
  volume = {381},
  number = {3},
  pages = {243--251},
  year = {2019},
  doi = {10.1056/NEJMoa1803180}
}

@article{magder2018meaning,
  author = {Magder, S.},
  title = {The meaning of blood pressure},
  journal = {Critical Care},
  volume = {22},
  number = {1},
  pages = {257},
  year = {2018},
  doi = {10.1186/s13054-018-2171-1}
}

@ARTICLE{pulsedb_paper,
AUTHOR={Wang, Weinan  and Mohseni, Pedram  and Kilgore, Kevin L.  and Najafizadeh, Laleh },
TITLE={PulseDB: A large, cleaned dataset based on MIMIC-III and VitalDB for benchmarking cuff-less blood pressure estimation methods},
JOURNAL={Frontiers in Digital Health},
VOLUME={Volume 4 - 2022},
YEAR={2023},
URL={https://www.frontiersin.org/journals/digital-health/articles/10.3389/fdgth.2022.1090854},
DOI={10.3389/fdgth.2022.1090854},
ISSN={2673-253X}}

@article{lee2022vitaldb,
  title={VitalDB, a high-fidelity multi-parameter vital signs database in surgical patients},
  author={Lee, Hyung-Chul and Park, Yoonsang and Yoon, Soo Bin and Yang, Seong Mi and Park, Dongnyeok and Jung, Chul-Woo},
  journal={Scientific Data},
  volume={9},
  number={1},
  pages={279},
  year={2022},
  publisher={Nature Publishing Group UK London}
}

@article{physionet,
author = {Ary L. Goldberger  and Luis A. N. Amaral  and Leon Glass  and Jeffrey M. Hausdorff  and Plamen Ch. Ivanov  and Roger G. Mark  and Joseph E. Mietus  and George B. Moody  and Chung-Kang Peng  and H. Eugene Stanley },
title = {PhysioBank, PhysioToolkit, and PhysioNet  },
journal = {Circulation},
volume = {101},
number = {23},
pages = {e215-e220},
year = {2000},
doi = {10.1161/01.CIR.101.23.e215},
URL = {https://www.ahajournals.org/doi/abs/10.1161/01.CIR.101.23.e215},
eprint = {https://www.ahajournals.org/doi/pdf/10.1161/01.CIR.101.23.e215}}

@article{johnson2016mimic,
  title={MIMIC-III, a freely accessible critical care database},
  author={Johnson, Alistair EW and Pollard, Tom J and Shen, Lu and Lehman, Li-wei H and Feng, Mengling and Ghassemi, Mohammad and Moody, Benjamin and Szolovits, Peter and Anthony Celi, Leo and Mark, Roger G},
  journal={Scientific data},
  volume={3},
  number={1},
  pages={1--9},
  year={2016},
  publisher={Nature Publishing Group}
}

@article{huang2022mlp,
  title={MLP-BP: A novel framework for cuffless blood pressure measurement with PPG and ECG signals based on MLP-Mixer neural networks},
  author={Huang, Bin and Chen, Weihai and Lin, Chun-Liang and Juang, Chia-Feng and Wang, Jianhua},
  journal={Biomedical Signal Processing and Control},
  volume={73},
  pages={103404},
  year={2022},
  publisher={Elsevier}
}

@inproceedings{yan2019novel,
  title={Novel deep convolutional neural network for cuff-less blood pressure measurement using ECG and PPG signals},
  author={Yan, Cong and Li, Zhenqi and Zhao, Wei and Hu, Jing and Jia, Dongya and Wang, Hongmei and You, Tianyuan},
  booktitle={2019 41st Annual International Conference of the IEEE Engineering in Medicine and Biology Society (EMBC)},
  pages={1917--1920},
  year={2019},
  organization={IEEE}
}

@article{whelton,
author = {Paul K. Whelton  and Robert M. Carey  and Wilbert S. Aronow  and Donald E. Casey  and Karen J. Collins  and Cheryl Dennison Himmelfarb  and Sondra M. DePalma  and Samuel Gidding  and Kenneth A. Jamerson  and Daniel W. Jones  and Eric J. MacLaughlin  and Paul Muntner  and Bruce Ovbiagele  and Sidney C. Smith  and Crystal C. Spencer  and Randall S. Stafford  and Sandra J. Taler  and Randal J. Thomas  and Kim A. Williams  and Jeff D. Williamson  and Jackson T. Wright },
title = {2017 ACC/AHA/AAPA/ABC/ACPM/AGS/APhA/ASH/ASPC/NMA/PCNA Guideline for the Prevention, Detection, Evaluation, and Management of High Blood Pressure in Adults},
journal = {JACC},
volume = {71},
number = {19},
pages = {e127-e248},
year = {2018},
doi = {10.1016/j.jacc.2017.11.006},

URL = {https://www.jacc.org/doi/abs/10.1016/j.jacc.2017.11.006},
eprint = {https://www.jacc.org/doi/pdf/10.1016/j.jacc.2017.11.006}

}

@article{hardin,
author = {Hardin, Amy and Hackell, Jesse},
year = {2017},
month = {08},
pages = {e20172151},
title = {Age Limit of Pediatrics},
volume = {140},
journal = {Pediatrics},
doi = {10.1542/peds.2017-2151}
}

@incollection{ghimire2023geriatric,
  author = {Ghimire, K. and Dahal, R.},
  title = {Geriatric Care Special Needs Assessment},
  booktitle = {StatPearls [Internet]},
  publisher = {StatPearls Publishing},
  address = {Treasure Island (FL)},
  year = {2023},
  month = {Feb},
  note = {Updated 2023 Feb 20; In: StatPearls Publishing; 2025 Jan-},
  url = {https://www.ncbi.nlm.nih.gov/books/NBK570572/}
}

@misc{nih2025,
  author = {{National Institutes of Health}},
  title = {Age},
  year = {2025},
  url = {https://www.nih.gov/nih-style-guide/age},
  note = {Accessed: 2025-07-08}
}

@incollection{weir2023bmi,
  author = {Weir, C. B. and Jan, A.},
  title = {BMI Classification Percentile and Cut Off Points},
  booktitle = {StatPearls [Internet]},
  publisher = {StatPearls Publishing},
  address = {Treasure Island (FL)},
  year = {2023},
  month = {Jun},
  note = {Updated 2023 Jun 26; In: StatPearls Publishing; 2025 Jan-},
  url = {https://www.ncbi.nlm.nih.gov/books/NBK541070/}
}

@misc{cdc2024bmi,
  author = {{Centers for Disease Control and Prevention}},
  title = {Adult BMI Categories},
  year = {2024},
  url = {https://www.cdc.gov/bmi/adult-calculator/bmi-categories.html},
  note = {Accessed: 2025-07-08}
}

@article{liu2012university,
  title={University of Queensland vital signs dataset: Development of an accessible repository of anesthesia patient monitoring data for research},
  author={Liu, David and G{\"o}rges, Matthias and Jenkins, Simon A},
  journal={Anesthesia \& Analgesia},
  volume={114},
  number={3},
  pages={584--589},
  year={2012},
  publisher={LWW}
}

@article{goldberger2000physiobank,
  title={PhysioBank, PhysioToolkit, and PhysioNet: components of a new research resource for complex physiologic signals},
  author={Goldberger, Ary L and Amaral, Luis AN and Glass, Leon and Hausdorff, Jeffrey M and Ivanov, Plamen Ch and Mark, Roger G and Mietus, Joseph E and Moody, George B and Peng, Chung-Kang and Stanley, H Eugene},
  journal={circulation},
  volume={101},
  number={23},
  pages={e215--e220},
  year={2000},
  publisher={Lippincott Williams \& Wilkins}
}

@inproceedings{caruna1993multitask,
  title={Multitask learning: A knowledge-based source of inductive bias},
  author={Caruna, Rich},
  booktitle={Machine learning: Proceedings of the tenth international conference},
  pages={41--48},
  year={1993}
}

@ARTICLE{lenet_paper,
  author={Lecun, Y. and Bottou, L. and Bengio, Y. and Haffner, P.},
  journal={Proceedings of the IEEE}, 
  title={Gradient-based learning applied to document recognition}, 
  year={1998},
  volume={86},
  number={11},
  pages={2278-2324},
  doi={10.1109/5.726791}
}

@misc{he2015deepresiduallearningimage,
      title={Deep Residual Learning for Image Recognition}, 
      author={Kaiming He and Xiangyu Zhang and Shaoqing Ren and Jian Sun},
      year={2015},
      eprint={1512.03385},
      archivePrefix={arXiv},
      primaryClass={cs.CV},
      url={https://arxiv.org/abs/1512.03385}, 
}

@inproceedings{Szegedy2016,
  author = {Christian Szegedy and Vincent Vanhoucke and Sergey Ioffe and Jon Shlens and Zbigniew Wojna},
  title = {Rethinking the Inception Architecture for Computer Vision},
  booktitle = {2016 IEEE Conference on Computer Vision and Pattern Recognition (CVPR)},
  year = {2016}
}

@inproceedings{Gu2022,
  author = {Albert Gu and Karan Goel and Christopher R\'e},
  title = {Efficiently Modeling Long Sequences with Structured State Spaces},
  booktitle = {International Conference on Learning Representations (ICLR)},
  year = {2022}
}

@article{Kohn2015,
  author = {Kohn, Julie C. and Lampi, Marsha C. and Reinhart-King, Cynthia A.},
  title = {Age-related vascular stiffening: causes and consequences},
  journal = {Frontiers in Genetics},
  year = {2015},
  volume = {6},
  pages = {112},
  doi = {10.3389/fgene.2015.00112},
  pmid = {25926844},
  pmcid = {PMC4396535},
  note = {eCollection 2015}
}

@ARTICLE{evans_bmi,
  
AUTHOR={Evans, Joyce M.  and Wang, Siqi  and Greb, Christopher  and Kostas, Vladimir  and Knapp, Charles F.  and Zhang, Qingguang  and Roemmele, Eric S.  and Stenger, Michael B.  and Randall, David C. },
         
TITLE={Body Size Predicts Cardiac and Vascular Resistance Effects on Men's and Women's Blood Pressure},
        
JOURNAL={Frontiers in Physiology},
        
VOLUME={Volume 8 - 2017},

YEAR={2017},

URL={https://www.frontiersin.org/journals/physiology/articles/10.3389/fphys.2017.00561},

DOI={10.3389/fphys.2017.00561},

ISSN={1664-042X}
}



\end{document}